\documentclass[smallextended]{svjour3}       
\usepackage{eso-pic}
\newcommand\AtPageUpperMyright[1]{\AtPageUpperLeft{%
 \put(\LenToUnit{0.5\paperwidth},\LenToUnit{-1cm}){%
     \parbox{0.5\textwidth}{\raggedleft\fontsize{9}{11}\selectfont #1}}%
 }}%
\newcommand{\conf}[1]{%
\AddToShipoutPictureBG*{%
\AtPageUpperMyright{#1}
}
}
\usepackage{graphicx,amsmath,geometry,subcaption,hyperref}
 \usepackage[ruled,vlined]{algorithm2e}
\usepackage{algorithmic,latexsym}
\graphicspath{{./figures/}}
\begin{document}
\title{Vision based Target Interception using Aerial Manipulation}
\subtitle{}
\author{ Lima Agnel Tony$^{*}$ \and Shuvrangshu Jana$^{*}$ \and Aashay Bhise$^{*}$ \and Varun V P$^{\dagger}$ \and Aruul Mozhi Varman S$^{\dagger}$\and Vidyadhara B V$^{*}$ \and Mohitvishnu S Gadde$^{*}$\and  Debasish Ghose$^{*}$  \and Raghu Krishnapuram$^{\dagger}$}
\institute{ $^*$Guidance Control and Decision Systems Laboratory \\
              Department of Aerospace Engineering\\
              Indian institute of Science\\
              Bangalore-12, India\\           
           \and
             $\dagger$Robert Bosch Center for Cyber-Physical Systems\\
             Indian Institute of Science\\
             Bangalore-12, India\\
}
\date{}
\maketitle
\conf{MBZIRC Symposium\\ADNEC, Abu Dhabi, 26-27 February 2020}
\vspace{-1.5cm}
\begin{abstract}
 Selective interception of objects in unknown environment autonomously by UAVs is an interesting problem. In this work, vision based interception is carried out. This problem is a part of challenge 1 of Mohammed Bin Zayed International Robotic Challenge, 2020, where, balloons are kept at five random locations for the UAVs to autonomously explore, detect, approach and intercept. The problem requires a different formulation to execute compared to the normal interception problems in literature. This work details the different aspect of this problem from vision to manipulator design. The frame work is implemented on hardware using Robot Operating System (ROS) communication architecture.
\end{abstract}

\section{Introduction}
\label{intro}
Unmanned systems underwent major changes in the past few decades. Technological advancements have contributed to its improved applicability which includes agriculture, monitoring, search and rescue, etc. to name a few. Each of these require a different capability in terms of hardware and software. In order to make the UAVs do some task, strategies need to be devised that break down the problem into sub tasks which in turn require individual formulation. In this work, such an application is addressed wherein, the UAV intercepts a specified target which is achieved in phases and with the aid of vision. The term interception could be inferred in various contexts, from counter UAV applications to fruit picking. A generalised method is conceived in our approach which could be employed for any of these applications. The phases of this approach could be roughly divided into target search, optimal approaching and tracking and interception.

UAVs are employed for wide variety of search applications. In \cite{r1},\cite{r1a}, UAVs are used to carry out inspection in bridges and infrastructures. This includes visual data collection followed by image processing to detect and then locate cracks. Similar application is seen in \cite{r1b}-\cite{r2} where, a UAV is assigned to perform power line surveillance for improving the performance and to carry out maintenance on time. 

Aerial vehicles are automated for various applications like package delivery, remote sensing, precision agriculture \cite{r3}, target tracking \cite{r4},  disaster management \cite{r6} etc., to name a few. Multi-drone coordination is an important aspect which is much useful in this regard. UAVs could coordinate in better achieving many of the tasks. One such application is interception or capture of objects. From harvesting of fruits in large farms to anti-drone systems in defense, this concept could be effectively deployed. In this work, we address the problem of interception of balloons in an area, as specified by the problem statement of challenge 1 of Mohammed Bin Zayed International Robotic Challenge 2020. The solution is developed in Robot Operating System (ROS) and further implemented on hardware.
The rest of the paper is organised as follows: Section \ref{2} gives the problem statement and Section \ref{3} details the methodology involved and details of implementations.  Section \ref{5} concludes the paper.

\section{Problem description}\label{2}
The challenge 1 arena for MBZIRC is 100 m x 40 m x 20 m in which there will be multiple randomly distributed balloons within a height of 5 m. These are mounted on 2 m pole with suitable mounting. The balloons are approximately 45 cm in diameter and are white in color. They are stationary but swaying slightly due to external wind disturbance. The objective is to pop the balloons. This would involve search, detect, approach and intercept the balloons recursively. 

\section{Formulation and implementation}\label{3}
\begin{figure}
    \centering
\begin{subfigure}{0.33\textwidth}
\includegraphics[width=1\linewidth, height=2.5cm]{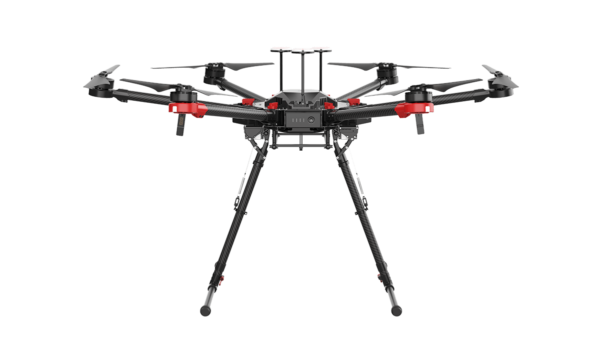} 
\includegraphics[width=1\linewidth, height=2.5cm]{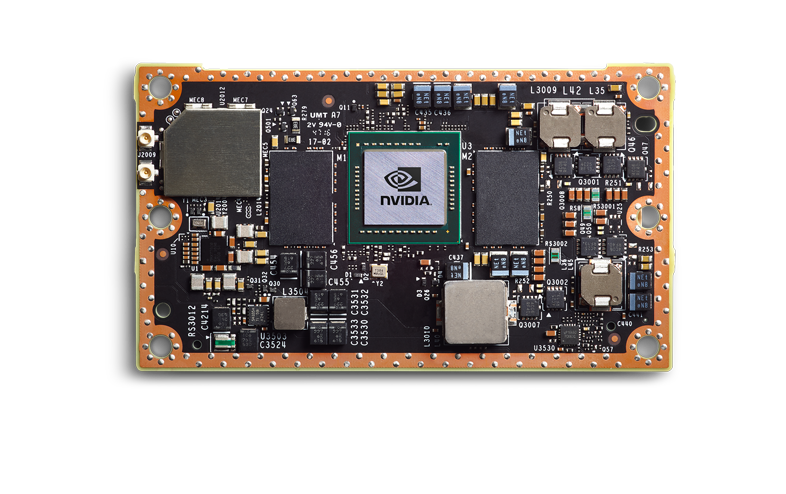}
\caption{(a)}
\end{subfigure}
\begin{subfigure}{0.3\textwidth}
\includegraphics[width=1\linewidth, height=2.5cm]{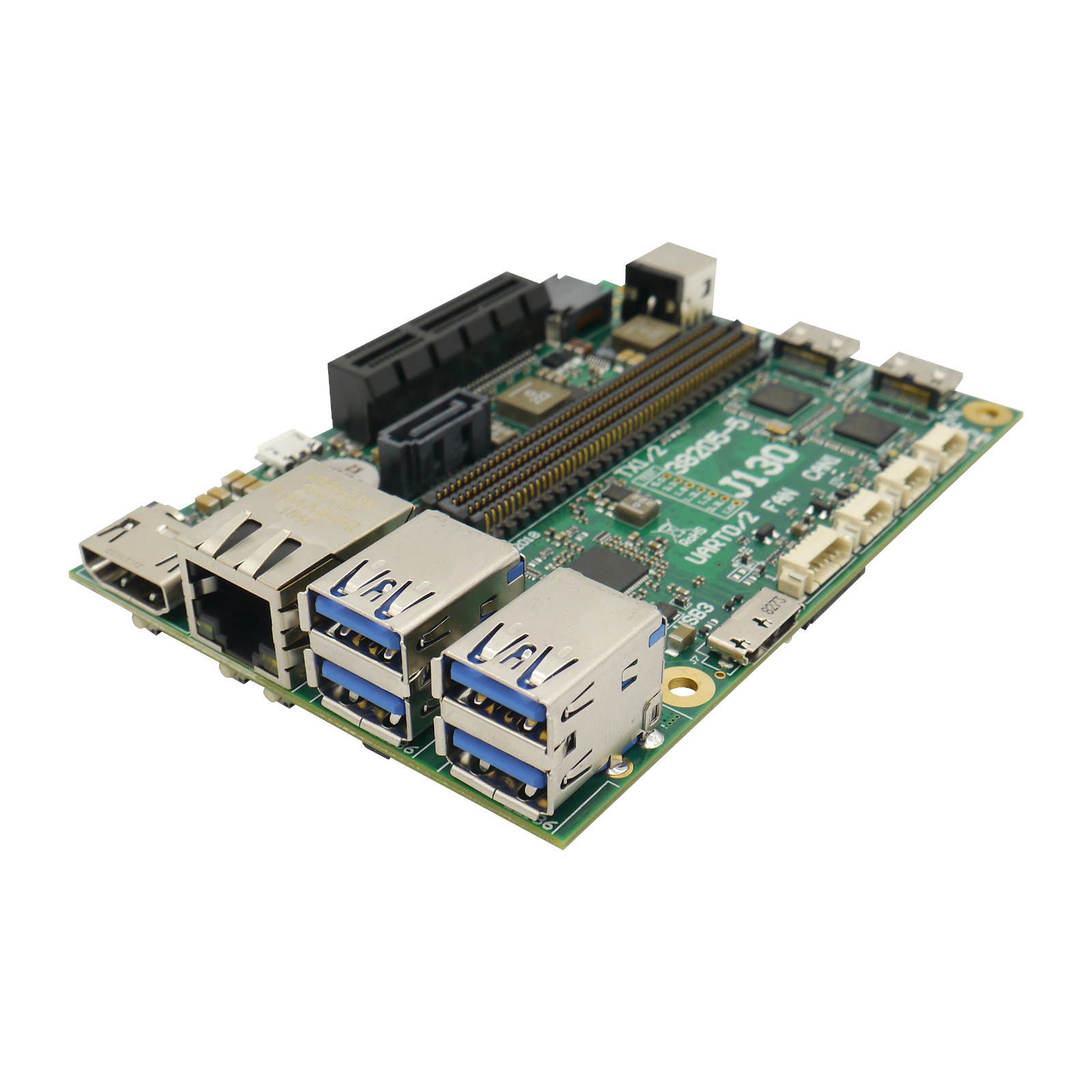} 
\includegraphics[width=0.5\linewidth, height=1.75cm]{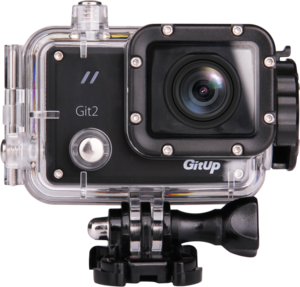}
\caption{(b)}
\end{subfigure}
\begin{subfigure}{0.3\textwidth}
\includegraphics[width=0.8\linewidth, height=2.25cm]{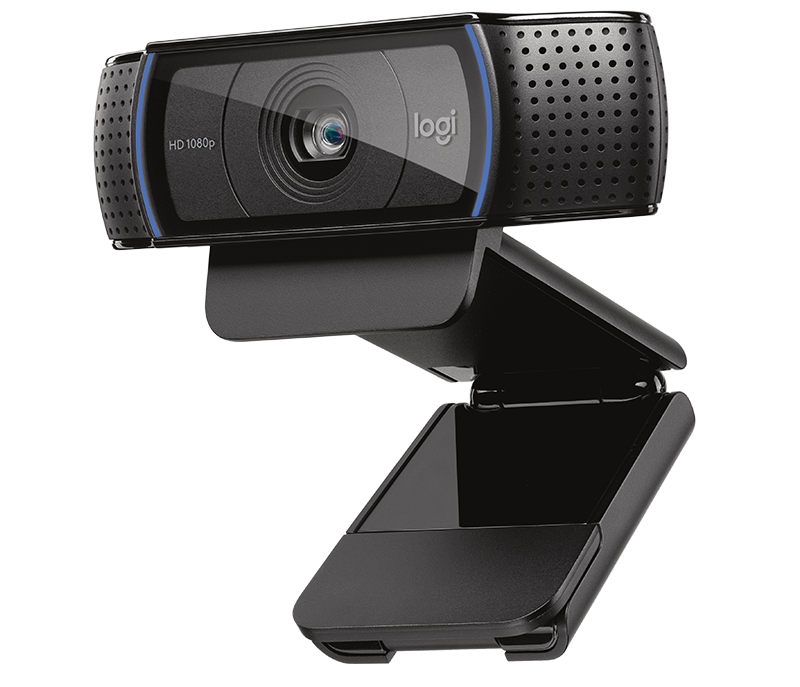}
\includegraphics[width=1\linewidth, height=2.25cm]{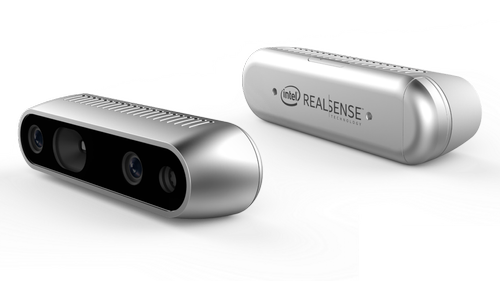}
\caption{(c)}
\end{subfigure}
    \caption{(a) DJI M600 pro and Nvidia Jetson TX2 (b) J130 carrier board and Gitup Git2P action camera (c) Logitech C920 HD web camera and Intel RealSense D435i}
    \label{f1}
\end{figure}

The algorithms are coded in C/C++ and vision modules are coded in Python. The hardware implementation is done using DJI M600 Pro hexacopter, as in Fig \ref{f1}(a). The UAV is equipped with the NVIDIA Jetson TX2 with Auvidea J130 carrier board as the companion computer. DJI OSDK ROS package is used to communicate with the A3 Pro autopilot. The DJI M600 Pro comes off the shelf with triple GPS redundancy. The localisation is achieved by global navigation satellite system (GNSS) fused with the onboard IMU. The drone is equipped with Tarot GOPRO 3D V metal 3-axis gimbal housing the Gitup Git2P action camera, Fig. \ref{f1}(b). The wide Field of View (FoV) of the camera facilitates the detection of the far away targets. A secondary camera, Logitech C920 web cam, is mounted on the manipulator body which aids the popping of the balloon in the terminal phase and Intel RealSense D435i is also equipped on the system for getting depth information (Fig. \ref{f1}(c)).
As mentioned in previous section, the problem could be sub divided into the following sub problems which are detailed below.

\subsection{Search}
\begin{figure}
    \centering
\begin{subfigure}{0.45\textwidth}
\includegraphics[width=0.9\linewidth, height=3.5cm]{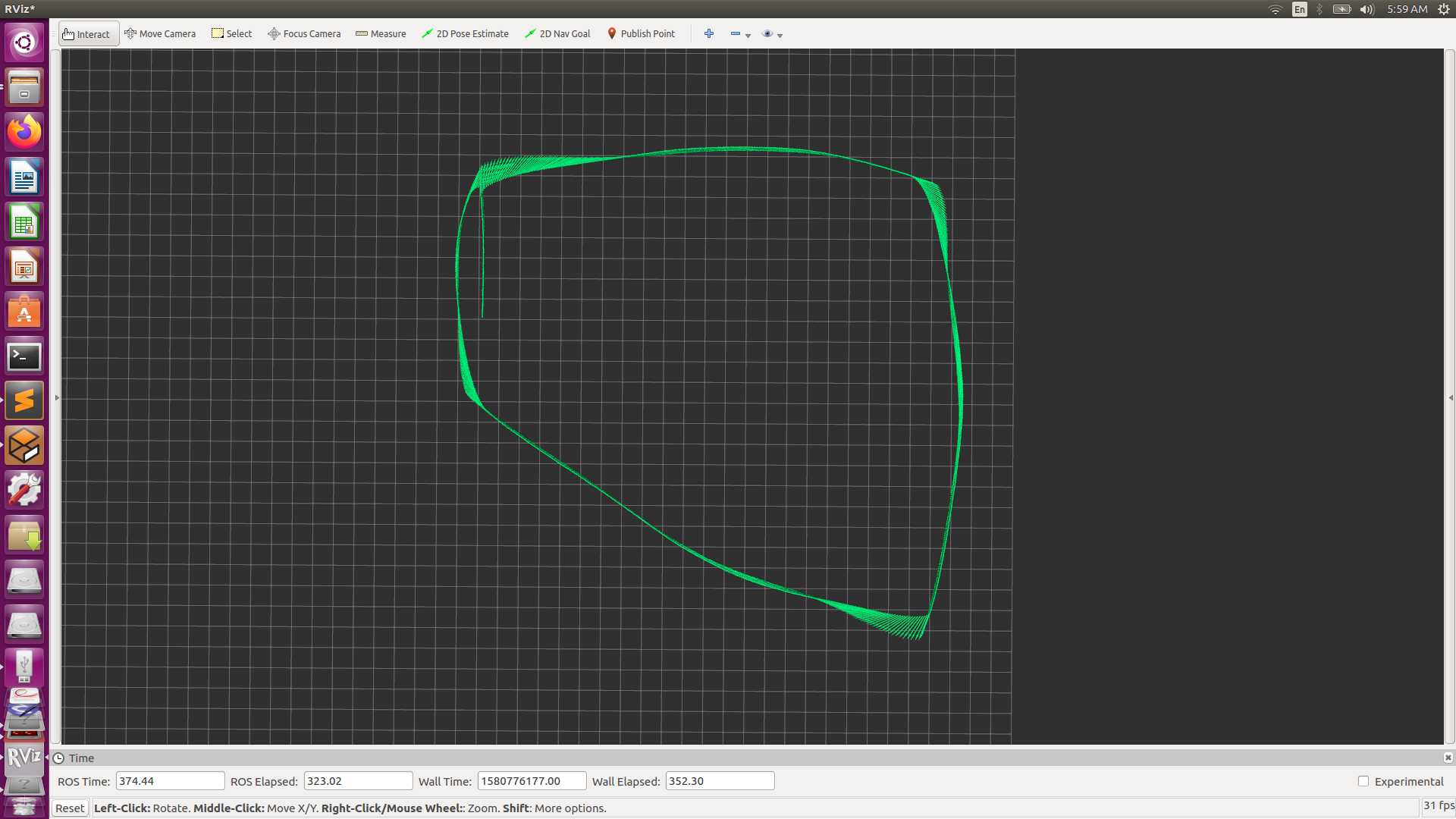}
\caption{(a)}
\end{subfigure}
\begin{subfigure}{0.45\textwidth}
\includegraphics[width=0.9\linewidth, height=3.5cm]{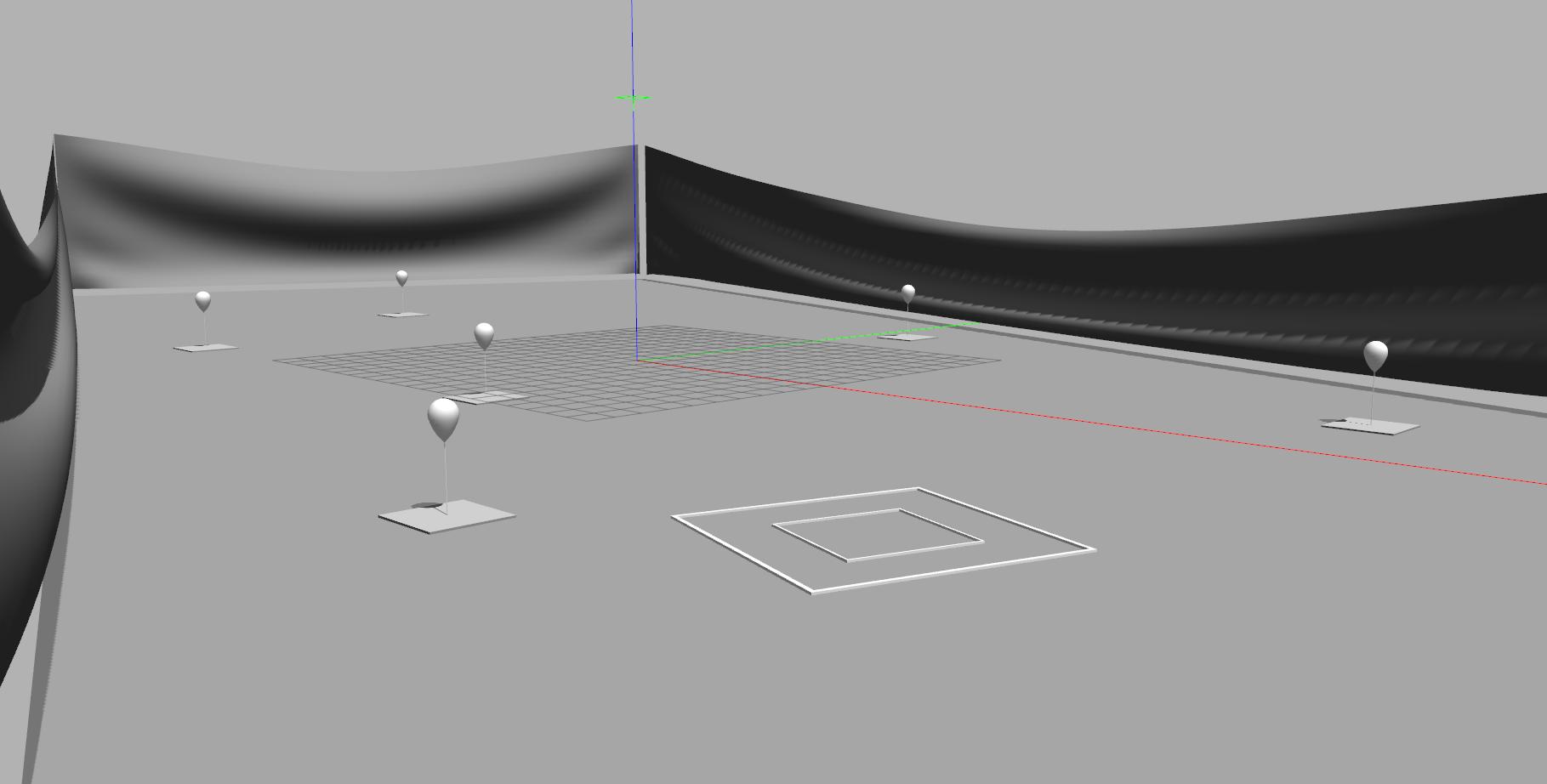} 
\caption{(b)}
\end{subfigure}
\caption{(a) Path employed for search operation (b) Arena set up in Gazebo}
\label{f2}
\end{figure}

The balloons are randomly located in an area of 100 m x 40 m and the search volume is restricted below 5 m. Considering the boundaries of the arena being covered by net, the effective search volume reduces to 90 m x 30 m x 5 m. A square search pattern is employed to cover the entire arena, as shown in Fig. \ref{f2}(a) (visualised in RViz). The UAV uses a forward facing camera to explore the volume while traversing through the path. The arena set up created in Gazebo environment is shown in Fig. \ref{f2}(b). The balloons are erected and their motion is simulated by adding wind effects to the environment.
The UAV takes off from a predefined starting location. It then moves to the starting point of the lawn mover pattern which is 4 m high. The choice of altitude is to aid the detection of balloons. One complete traversal takes place when the vehicle starts from point A and reaches point B. The same is implemented on DJI M600 hexacopter. The vehicle executes the search operation at a max speed of 2 m/s. Velocity commands are sent to the autopilot which are computed on the companion computer based on the search pattern. 
\subsection{Vision based balloon detection}
The vision module comprises of a deep neural network to detect the balloons and a series of trackers to track the detected balloons. Tiny YOLOv3\cite{r7}, a shallow version of the latest YOLO\cite{r8} object detection network trained on Pascal VOC\cite{r9} 2012 is re-purposed as a balloon detector by retraining the deeper layers. The detected balloons are tracked using Kalman filter in the spatial plane and Hungarian algorithm is used to associate the bounding box detections of the balloon detector with the trackers.
The Kalman filter is used to predict the bounding box of the tracked balloons and minimise the effect of inaccuracies in detections while tracking by modelling the state $x_n$ of the drones from the observations $z_1:n$. The spatial displacement of the balloons in the image plane are approximately modelled using constant velocity model, and the following states are used to track the balloons:
\begin{equation}
    x= [C_x, C_y, w, h]^T
\end{equation}
where $C_x$ and $C_y$ are $x$ axis coordinate and $y$ axis coordinate of the bounding box center. $W$ and $h$ are width and height of the bounding boxes respectively. 

\begin{figure}
    \centering
  \begin{subfigure}{0.45\textwidth}
\includegraphics[width=0.8\linewidth, height=2.5cm]{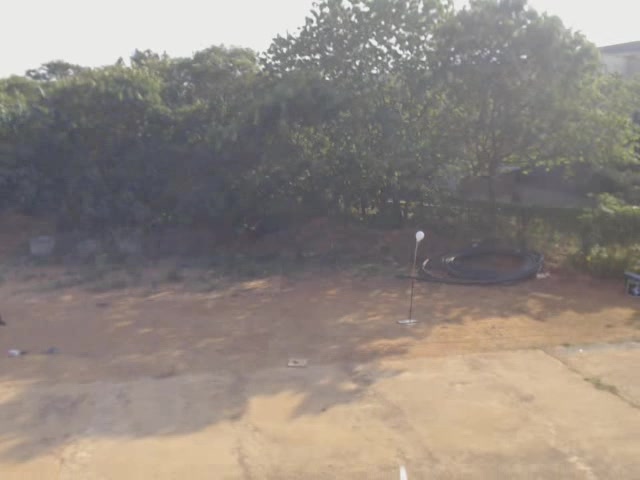}
\caption{}
\end{subfigure}
\begin{subfigure}{0.45\textwidth}
\includegraphics[width=0.8\linewidth, height=2.5cm]{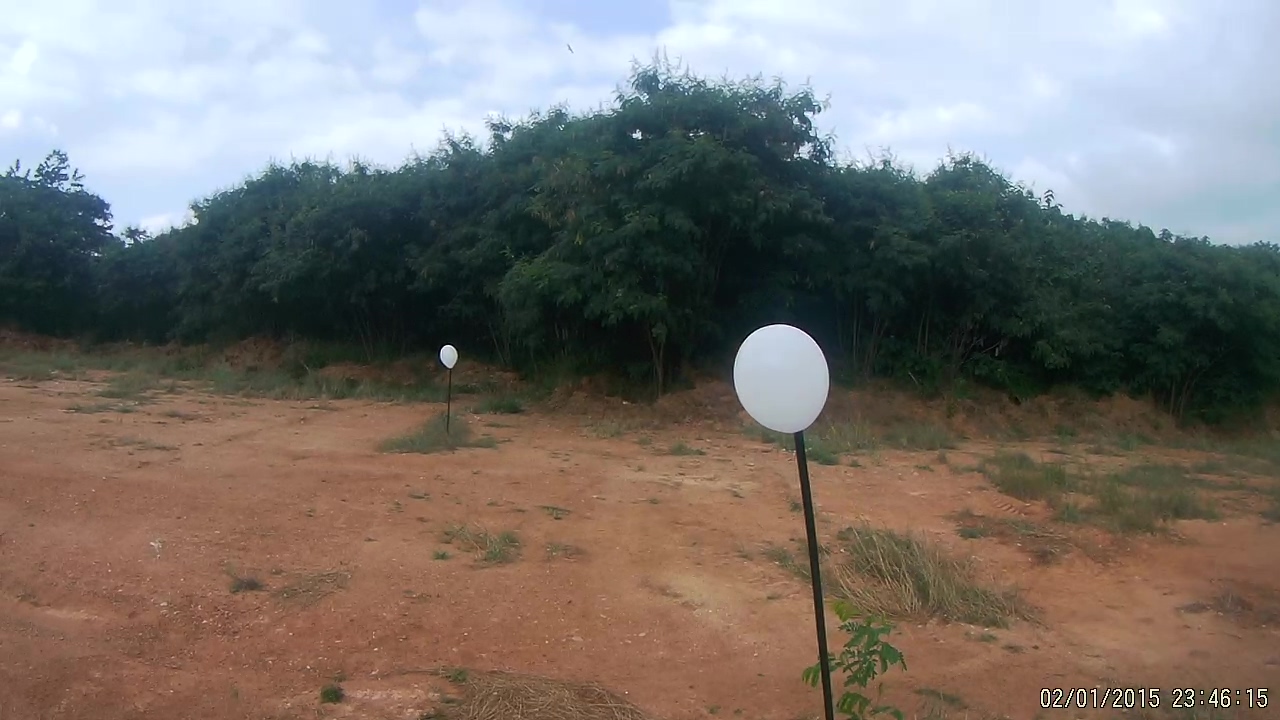}
\caption{}
\end{subfigure}
\begin{subfigure}{0.45\textwidth}
\includegraphics[width=0.8\linewidth, height=2.5cm]{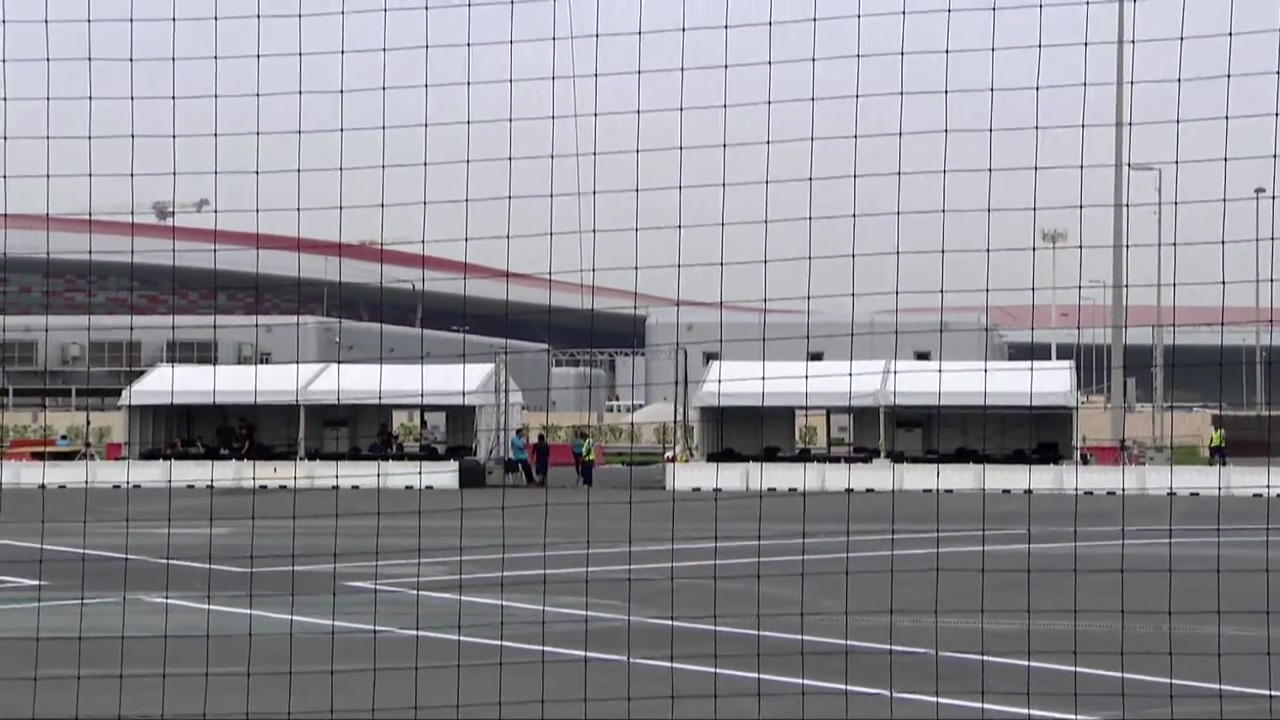}
\caption{}
\end{subfigure}
\begin{subfigure}{0.45\textwidth}
\includegraphics[width=0.8\linewidth, height=2.5cm]{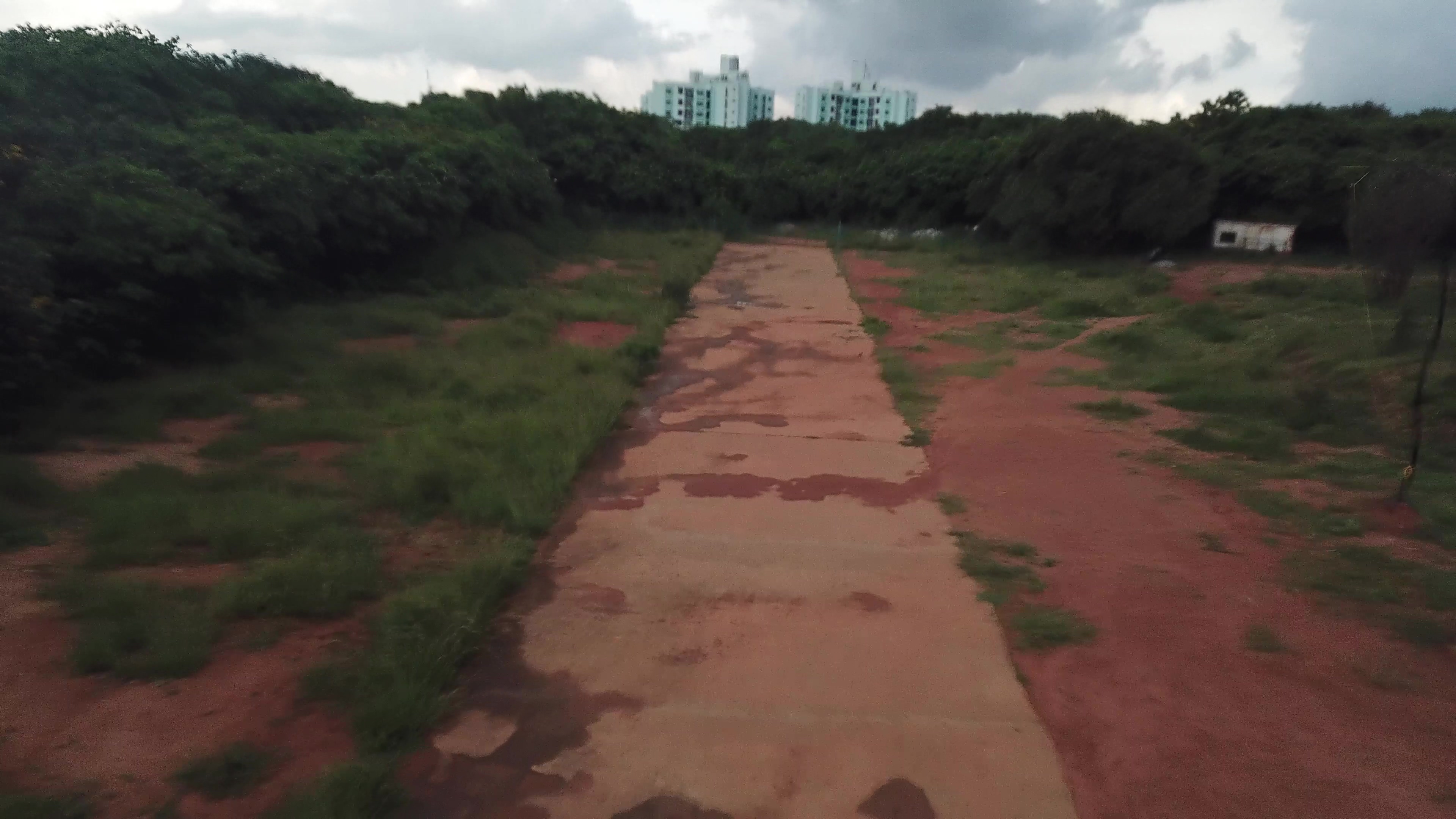}
\caption{}
\end{subfigure}
\caption{(a)-(b) Annotation data (c)-(d) Some negative samples}
    \label{f3a}
\end{figure}
Euclidean distance is used by the Hungarian algorithm as a metric to associate the bounding box detections of balloons from the deep neural network with the balloons tracked by the Kalman filters. The cost matrix is computed by calculating the Euclidean distance between the centers of the deep neural network detections and the Kalman filter predictions.
For every tracked balloon at each frame $n$, the bounding box prediction of the tracked balloon is corrected by updating the state of the trackers with the associated bounding box detections. A new Kalman filter instance with unique ID is initialized whenever a bounding box detection does not correspond to any of the existing tracked balloons, and the detection is used to initialize the state of the new balloon tracker. When no observations can be associated with a tracked balloon, its bounding boxes are predicted using linear velocity model without any corrections for the prediction at that frame. A balloon tracker is removed when bounding box detections cannot be associated with the tracked balloon for multiple successive frames.
The regions of the corrected bounding boxes corresponding to the balloons are segmented for the colour of the balloon and a circle is fit along the major axis for each of the balloons. The rough location of the balloons in the camera frame are calculated by assuming that the size of the balloons are known and they are spherical in shape.
\begin{figure}
    \centering
  \begin{subfigure}{0.45\textwidth}
\includegraphics[width=0.8\linewidth, height=2.5cm]{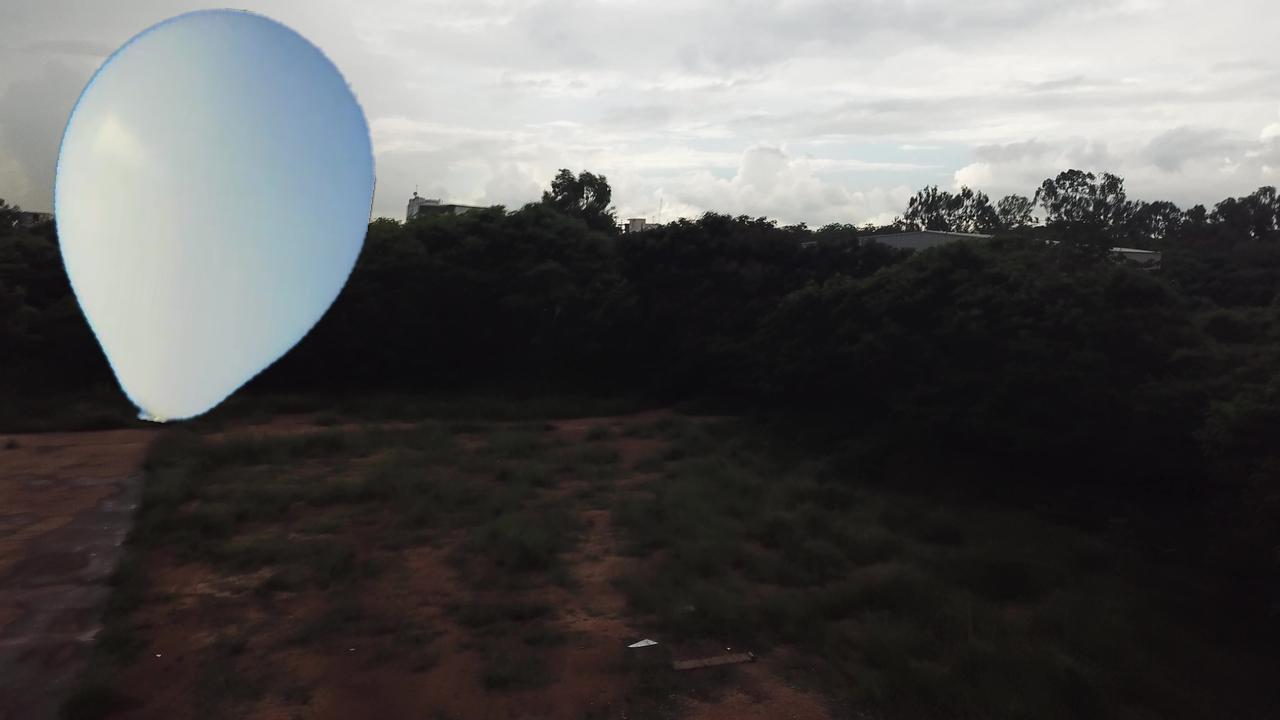}
\caption{}
\end{subfigure}
\begin{subfigure}{0.45\textwidth}
\includegraphics[width=0.8\linewidth, height=2.5cm]{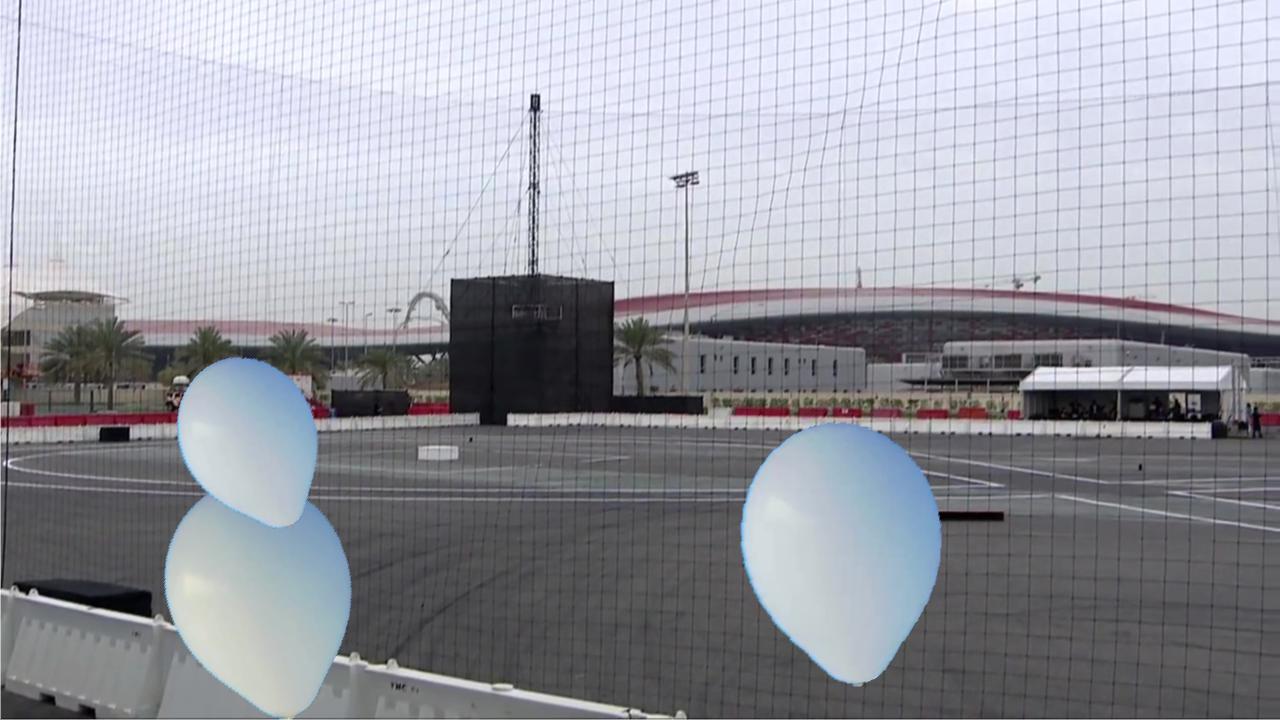}
\caption{}
\end{subfigure}
\begin{subfigure}{0.45\textwidth}
\includegraphics[width=0.8\linewidth, height=2.5cm]{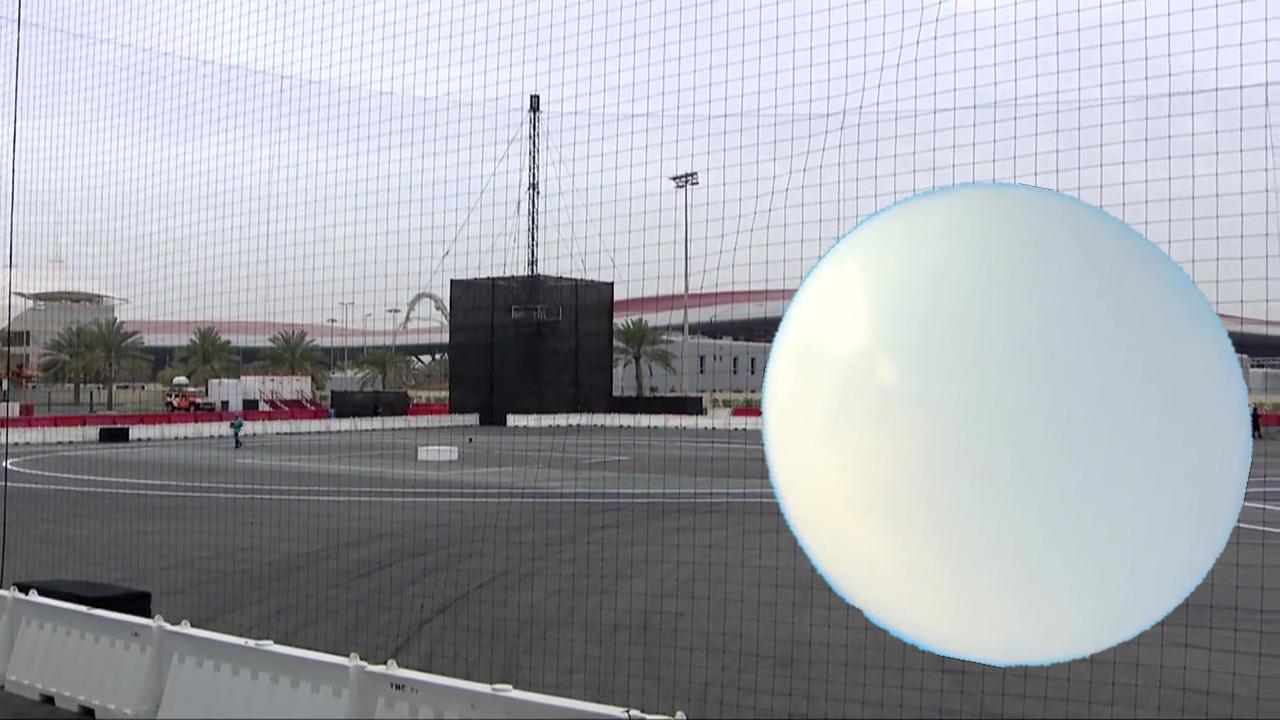}
\caption{}
\end{subfigure}
\begin{subfigure}{0.45\textwidth}
\includegraphics[width=0.8\linewidth, height=2.5cm]{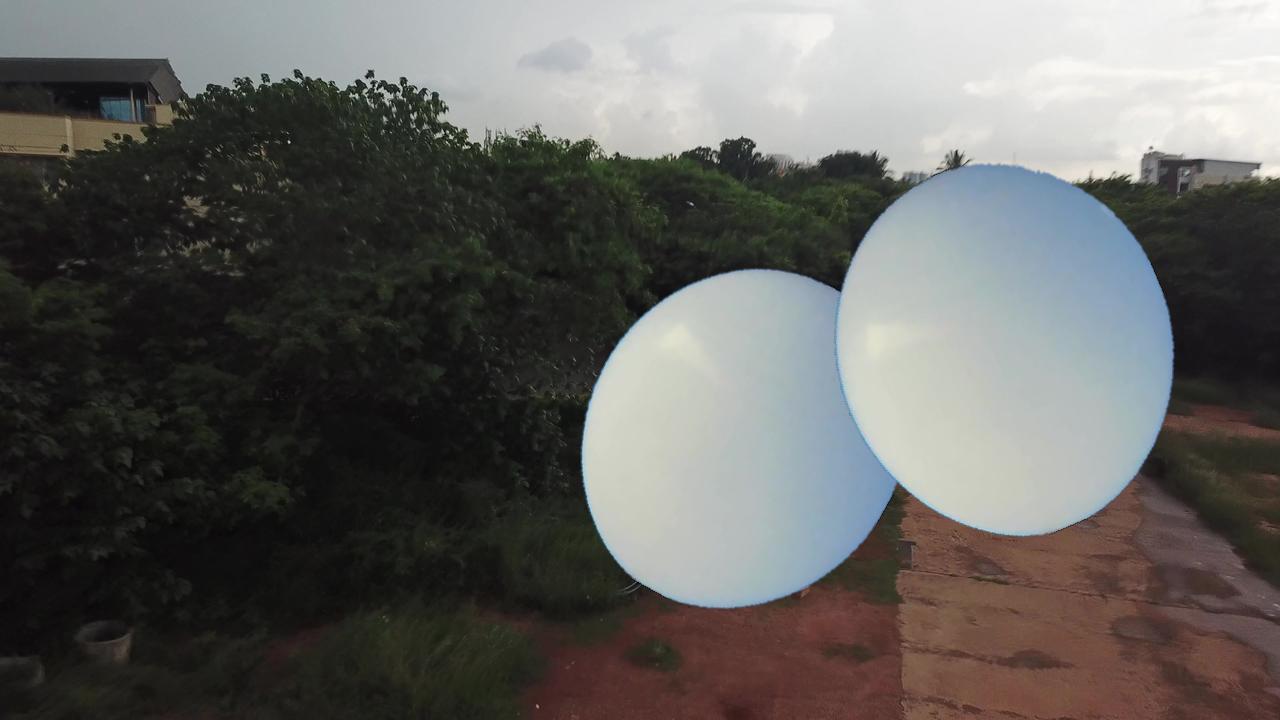}
\caption{}
\end{subfigure}
    \caption{(a)-(d) few samples of synthetic data generated }
    \label{f3b}
\end{figure}
TinyYOLOv3 was chosen to train the balloon detector as it’s smaller in size, computationally lighter and can be easily deployed on Nvidia Jetson TX2 mounted on the drone. YOLO approaches use a single neural network to predict the bounding boxes and the it’s class probabilities unlike other object detection networks based on region proposal networks like Faster R-CNN\cite{r10}.
TinyYOLOv3 trained on PASCAL VOC 2012 data set is re-purposed as a balloon detector by fine-tuning the last four convolutional layers of the network. Both manually annotated data and synthetically generated data set are used to re-train the network. A large portion of the data set is synthetically generated by blending the balloon images collected at different lighting conditions with diverse variety of backgrounds at different sizes and orientations as described in \cite{r11}. To extract the balloons for synthetic data generation Pixel Objectness\cite{r12}, a deep neural network for semantic segmentation is used to segment the balloons from the backgrounds. Also, many negative samples without any annotations are also included in the data set. The inclusion of synthetically generated images in the data set used for training the network improves the mAP@0.50 from 68 to 98. From the results, it can be observed that the inclusion of many negative samples and uniformly generated annotations of the synthetic data set improves the mAP of the detector. Darknet\cite{r13}, an open source neural network is used to train the deep neural network and the trained model is converted to a Tensor Flow model. The converted model is then deployed with INT8 precision on Nvidia Jetson TX2 using TensorRT optimized TensorFlow\cite{r14} for accelerated inference. The output of the module on a test data is given in Fig. \ref{f3b}.
\begin{figure}
    \centering
    \includegraphics[scale=0.5]{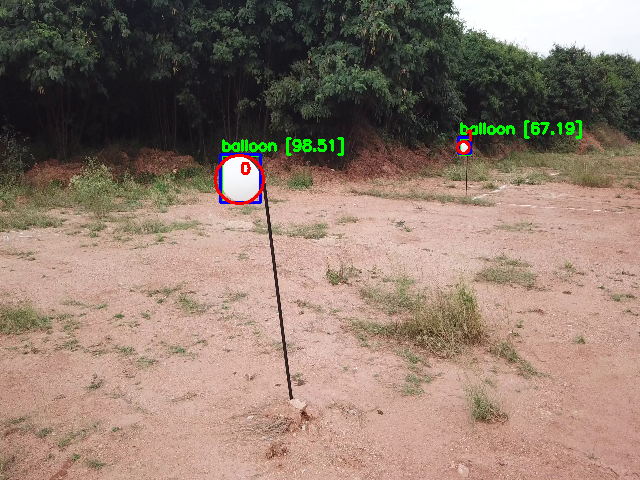}
    \caption{Output of the detection module}
    \label{f3b}
\end{figure}
\subsection{Approaching the swaying target}
 The  popping of balloon is considered as  interception of stationary target with an agent. Upon consecutive positives affirming the target is in FoV by the balloon detection module, the UAV corrects its heading towards the detected target and move towards the target. The drone is  subjected  to velocity and yaw rate commands so the popping mechanism engages with the balloon. If  $ p_{x}$ , $p_{y}$ are the  pixel location of  the  object,  then the unit vector towards the   direction of the object in camera frame is,
 \begin{equation}
  o^{c}= \frac{1}{\sqrt{p_{x}^{2}+p_{y}^2+f^{2}}} (p_{x}, p_{y}, f)
 \end{equation}
 where, f is the focal length.  If the required magnitude of the velocity is  $V $, the commanded velocity along the $x$, $y$, and $z$ axes of camera frame to intercept the object are
 \begin{eqnarray}
 v_{x}&=& V\frac{p_{x}}{\sqrt{p_{x}^{2}+p_{y}^2+f^{2}}}\\
v_{y}&=& V\frac{p_{y}}{\sqrt{p_{x}^{2}+p_{y}^2+f^{2}}}\\
v_{z}&= &V\frac{p_{z}}{\sqrt{p_{x}^{2}+p_{y}^2+f^{2}}}
 \end{eqnarray}
The desired yaw is calculated as
\begin{equation}
   \psi_{\text{des}}= \arctan(\frac{p_{y}}{p_{x}})
\end{equation}
The UAV is subjected to  velocities and yaw rate as follows:
\begin{equation}
    V_{f}= R_{1}R_{2}V_{c}
\end{equation}
where, $V_{f}$ and $V_{c}$ are the velocity vector in the vehicle frame  and camera frame, $R_{1}$ and $R_{2}$ are the rotation matrix from camera frame to body frame and body frame to vehicle frame respectively.

\subsection{Interception and confirmation}
The drone is equipped with sharp ended manipulator mechanism which aids in popping of the balloon.
The drone proceeds towards the detected balloon after aligning towards it. After some time, balloon would be completely out of FOV. This would mean that the vehicle has either popped the balloon or missed the target. Confirmation on popping is carried out by executing a revisit to the location at which the balloon was detected and searching for any balloon in FOV of the camera. If the vehicle fails to burst the balloon, repeated trials are carried out in the revisits.
In case of multiple balloons in FoV, an indexing is used to carry out successive interceptions. Balloons are numbered based on the depth. Intel RealSense D435i RGB-D camera is used to measure the separation between the vehicle and the balloons. 

\begin{figure}
    \centering
    \includegraphics[scale=0.265]{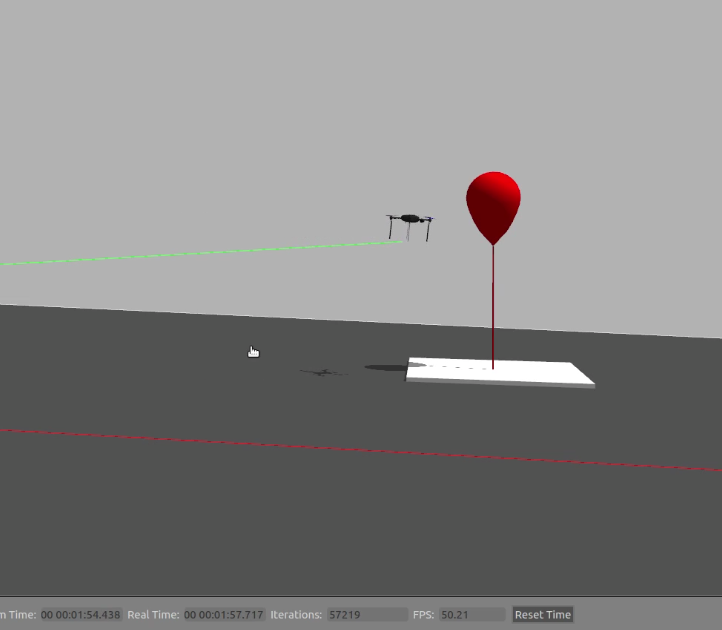}
    \includegraphics[scale=0.2]{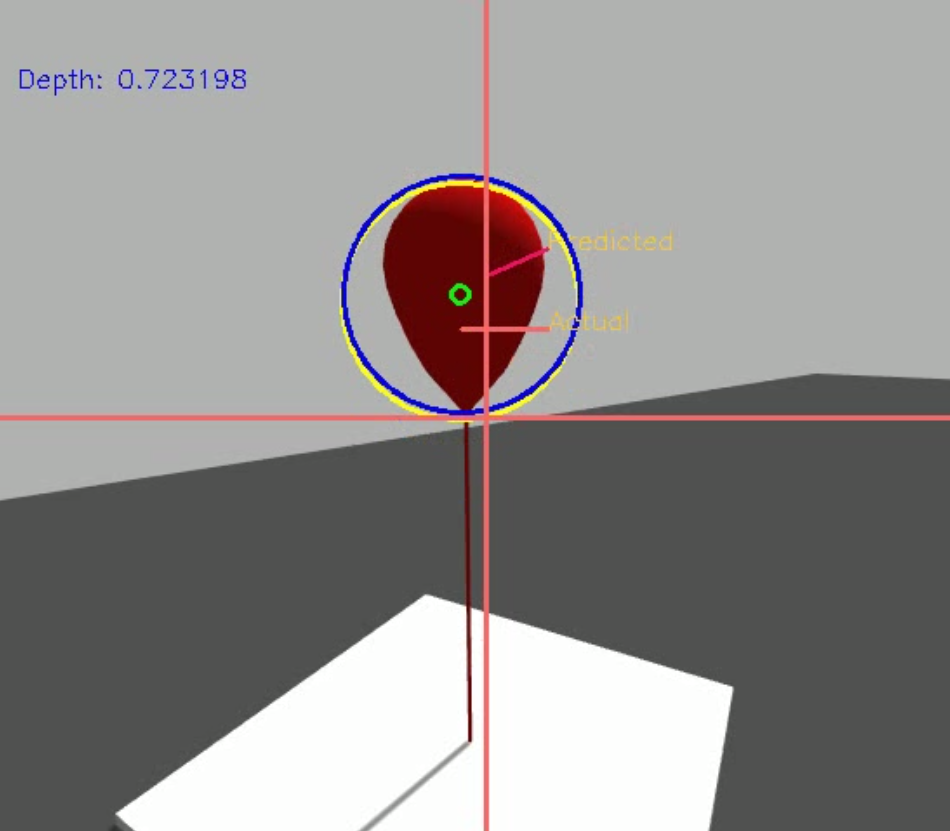}
    \caption{(a) Drone immediately before popping (b) Cam view}
    \label{fig}
\end{figure}

The simulation for the interception scenario, as visualised in Gazebo can be found in \footnote{\href{https://indianinstituteofscience-my.sharepoint.com/:f:/g/personal/limatony_iisc_ac_in/Ei-19wlyHY9GpEWuSTTBfscBK11dbYnB74v2J83yznHANg?e=ewZju7}{Gazebo simulation}}. The simulations contains the third eye as well as the drone manipulator camera FOV. As white balloon detection is not properly working, we have used red detection in this simulation (Fig. \ref{fig}).

The implementation is done using M600 pro drone and the specifically designed and prototyped manipulator arm. The video of popping could be found in\footnote{\href{https://indianinstituteofscience-my.sharepoint.com/:f:/g/personal/limatony_iisc_ac_in/Eri6fJjaga5EoEJ8MGz0Yq0BCTtBe52iQOAzYQeccrQZLg?e=dzkZKp}{Hardware implementation}}. It can be seen that the mechanism works precisely to pop the balloons in its FOV.
\subsection{Task management}\label{4}
A peripheral layer is developed which would coordinate multiple vehicles and perform task allocation for a multi-UAV interception. For partitioning the arena among the agents for performing interception, Voronoi partition method is used.  The outline of the balloon popping  algorithm is given in the following algorithm \ref{a1}.

\begin{algorithm}[H]
\SetAlgoLined
Assign area to drones \\
 Start exploration\\
   \eIf{Balloons in FoV}{
   change course to the detected region\\
   align towards the balloon\\
   move towards the center of the balloon\\
    \While{balloon in frame }{
    continue moving forward\\}
   }{
   Continue search in lawn mover pattern\\
  }
  \eIf{one agent fails}{
   Assign the  balance area to neighbour bot
  }
\protect\caption{Balloon popping using multiple UAVs}
\label{a1}
\end{algorithm}
The peripheral task manager not only assigns dedicated interception field for the UAVs but also makes sure that the UAVs don't hit each other by enabling inter agent collision avoidance. In the restricted space like the MBZIRC competition arena, it's very important to limit the motion of the vehicles within the volume. A geofencing sub module makes sure that this boundary condition is never compromised. This external manger has a very important role in affirming that multiple UAVs doesn't try to intercept on to the same balloon. This part of the work is an ongoing one and would be reported in subsequent publications.

\section{Conclusions}\label{5}
In this work, vision based interception of balloons by UAVs is being discussed. An RNN based vision module is employed to detect the balloons and avoid false positives. A specifically designed manipulator is fabricated to pop the balloon. The process involves exploration of the arena followed in a predetermined pattern until balloon is detected by the onboard camera. The UAV upon detection, orients the manipulator face towards the balloon. The UAV manipulator has sharp ended protrusions which aid in the effective popping of the balloons. The popping is confirmed by coming back and checking the FOV of the camera again. This can be practically used for several applications like fruit picking etc.
\begin{acknowledgements}
 We would like to acknowledge the Robert Bosch Center for Cyber Physical Systems, Indian Institute of Science, Bangalore, and Khalifa University, Abu Dhabi, for partial financial support. We would also like to thank fellow team members from IISc for their invaluable contributions towards this competition. 
\end{acknowledgements}

\end{document}